\documentclass[10pt,twocolumn,letterpaper]{article}

\usepackage{cvpr}
\usepackage{times}
\usepackage{epsfig}
\usepackage{graphicx}
\usepackage{amsmath}
\usepackage{amssymb}
\usepackage[stable]{footmisc}

\usepackage[breaklinks=true,bookmarks=false]{hyperref}

\cvprfinalcopy 

\def\cvprPaperID{****} 

\begin{document}

\title{The Herbarium Challenge 2019 Dataset}

\author{
Kiat Chuan Tan\thanks{Equal contribution.}\\
Google Research\\
{\tt\small kiatchuan@google.com}
\and
Yulong Liu$^*$\\
Google Research\\
{\tt\small liuyl@google.com}
\and
Barbara Ambrose\\
New York Botanical Garden\\
{\tt\small bambrose@nybg.org}
\and
Melissa Tulig\\
New York Botanical Garden\\
{\tt\small mtulig@nybg.org}
\and
Serge Belongie\\
Google Research \& Cornell Tech\\
{\tt\small sjb344@cornell.edu}
}
\maketitle

\begin{abstract}
Herbarium sheets are invaluable for botanical research, and considerable time and effort is spent by experts to label and identify specimens on them. In view of recent advances in computer vision and deep learning, developing an automated approach to help experts identify specimens could significantly accelerate research in this area. Whereas most existing botanical datasets comprise photos of specimens in the wild, herbarium sheets exhibit dried specimens, which poses new challenges. We present a challenge dataset of herbarium sheet images labeled by experts, with the intent of facilitating the development of automated identification techniques for this challenging scenario.
\end{abstract}

\section{Introduction}

We are currently in the midst of the sixth great extinction, Anthropocene extinction, that is largely being caused by humans. A recent summary of a forthcoming United Nations report indicates that 1,000,000 species could go extinct within decades \cite{un-ipbes-report}. As large ecosystems are being destroyed, it is likely that many of these extinctions are undocumented as many species still await discovery. It cannot be overstated that plants are fundamental to life on earth; they convert energy into food for all organisms, produce oxygen needed by most organisms, provide shelter, medicines, fuel, hold soil in place, and help regulate the water cycle. The loss of plant species is increasing the fragility of ecosystems which intensifies the devastating effects of fires and flood.  Plant species loss may be removing potentially valuable resources for human health and well-being before we have the opportunity to study and cultivate them. 

There are currently 400,000 known plant species and there are an estimated 80,000 plant species still to be discovered. Without new tools for species discovery, it is likely that we will lose these species to the current extinction event before we know their names, specific benefits, and the role they keep in maintaining the delicate balance of the ecosystems in which they occur. The major challenge to understanding plant diversity is speeding up the process of discovery. In flowering plants, it takes an average of 35 years from plant collection to species description while less than 16\% of new species are described in less than 5 years \cite{Bebber22169}. It has also been suggested that ‘Herbaria are a major frontier for species discovery’ with more than 50\% of unknown species already in herbarium collections \cite{Bebber22169}.

\begin{figure}[t]
\begin{center}
   \includegraphics[width=0.8\linewidth]{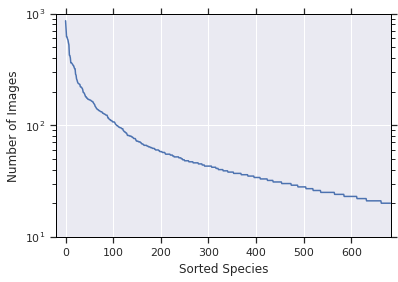}
\end{center}
   \caption{Number of imaged sheets per species, sorted from most common to least common.}
\label{fig:species_distribution}
\end{figure}

The use of Artificial Intelligence for automatic species identification is a new and rapidly developing field and holds promise for identifying the still to be named species in herbaria \cite{Carranza-Rojas2017,Clark2012,Soon98,Schuettpelz2017, Unger2016,Waldchen2018,Wijesingha2012,Wilf3305,Younis2018}. Several studies of automated plant species identification studies have focused on images of leaves alone \cite{Soon98,Wilf3305} while others have performed automatic species identification using herbarium specimens trained with leaf images \cite{Carranza-Rojas2017,Wijesingha2012}. Several studies have been performed using herbarium images alone \cite{Carranza-Rojas2017,Clark2012,Mabberley2017,Schuettpelz2017,Unger2016,Younis2018}, however, new models are still needed to automatically identify species using herbarium specimens alone and under conditions that more accurately reflect the distribution of specimen images available for species present in herbaria.

There are approximately 3,000 herbaria world-wide with an estimated 380 million specimens. The New York Botanical Garden (NYBG) has more than 7.8 million herbarium specimens, and over 3 million of these have been imaged. Each specimen is accompanied by an approximate standard set of metadata, including the name of the plant, a description of the location where it was collected, name of the collector, the accession number of the collector, and the date of collection.

\section{The Herbarium Dataset\footnote{Available at https://github.com/visipedia/herbarium\_comp}}
The Herbarium dataset contains 46,469 digitally imaged herbarium sheets representing 683 species from the flowering plant family Melastomataceae (melastome). This family is also known as the princess flower family and many species have extraordinary flowers. The Melastomataceae is a large family with more than 160 genera and more than 5,500 named species \cite{Mabberley2017}. Each herbarium specimen was carefully identified and labeled with a unique species by human experts from the New York Botanical Garden. We have maintained the original high resolution images as digitally imaged by the New York Botanical Garden.

Some melastome species are represented by more than a hundred specimens while others are represented by less than 20 specimens, which more accurately reflects the real-world distribution of specimens per species found in herbaria. See figure \ref{fig:species_distribution} for a distribution of the counts per species.

We calculated the overlap between species contained in the Herbarium challenge dataset with the plant species in the iNaturalist 2018 challenge dataset \cite{DBLP:journals/corr/HornASSAPB17}. The overlap only comprises 2 out of the 683 species in the Herbarium dataset.

\subsection{Dataset Challenges}
\begin{figure}[t]
\begin{center}
   \includegraphics[width=1.0\linewidth]{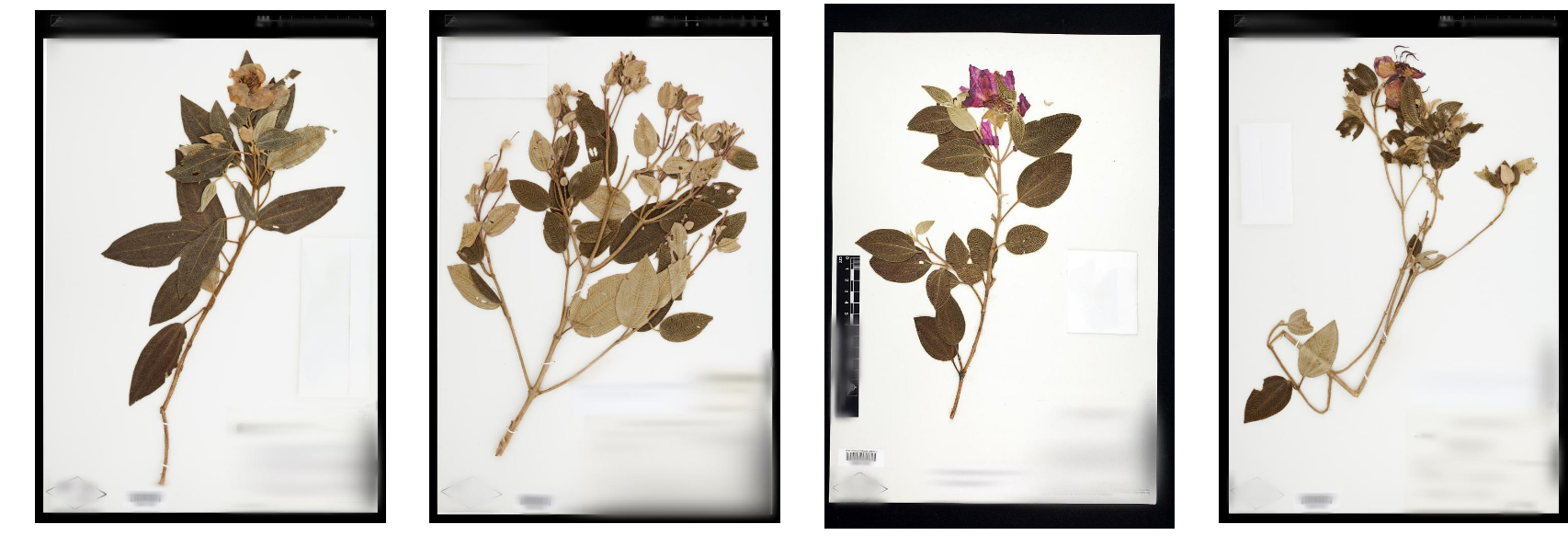}
\end{center}
   \caption{Different specimens of the same species, varying by specimen age.}
\label{fig:different_age}
\end{figure}

\begin{figure}[t]
\begin{center}
   \includegraphics[width=1.0\linewidth]{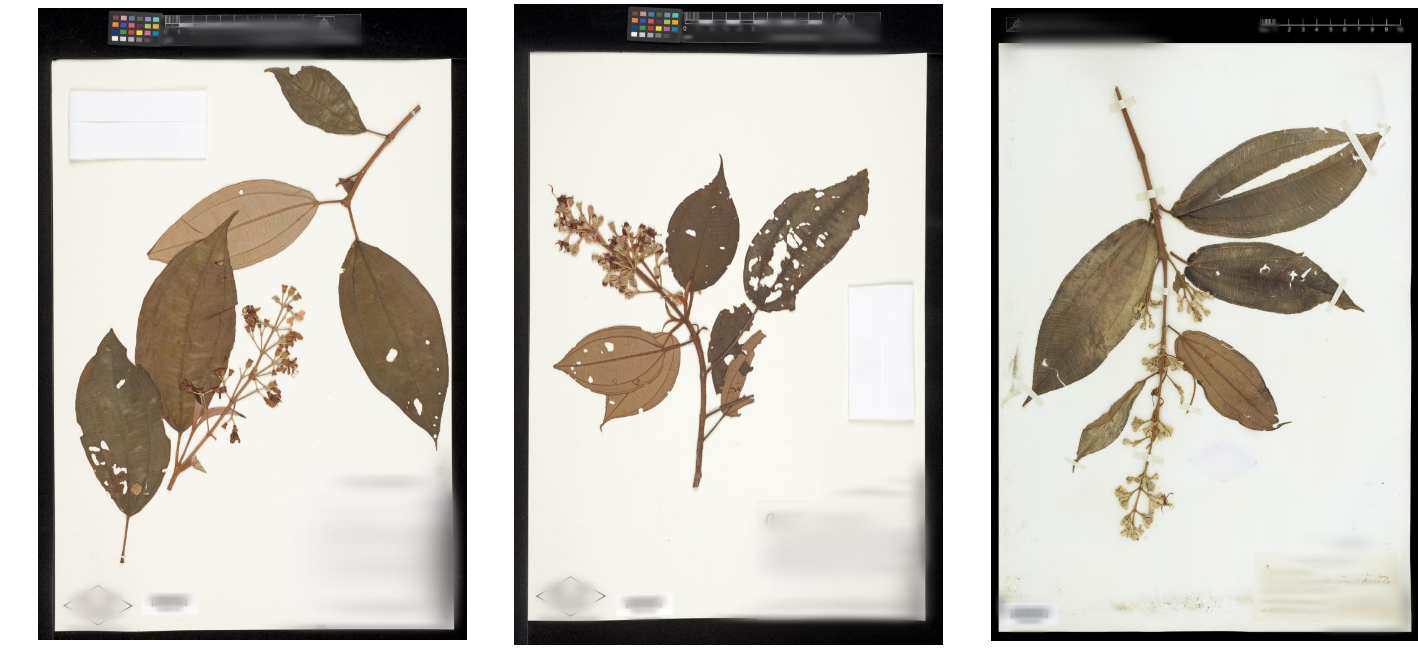}
\end{center}
   \caption{Visually similar specimens of 3 different species.}
\label{fig:similar_species}
\end{figure}

The Herbarium dataset presents multiple challenges for species identification. First, the dataset has a large class imbalance. Second, expert botanists often rely on characters that occupy a small portion of the herbarium sheet in order to identify the species, such as leaf veins, leaf margins, and flower shape \cite{COPE20127562}\cite{Waldchen2018}. Most of the existing state-of-the-art image classification methods use a low resolution image as input \cite{inception-v4}. This lower resolution may be insufficient to capture these smaller characters, and existing networks may find it challenging to learn to focus on these small differences. Third, individuals of the same species can vary widely in their morphology, i.e. intraclass variation can be large. For example, a specimen's color composition can change with age (figure \ref{fig:different_age}). Finally, the interclass variation can be small, i.e. different species can be visually very similar (figure \ref{fig:similar_species}).

\subsection{Dataset Split}
We split the dataset into 75\% training, 5\% validation and 20\% test. The split was performed randomly on a per-species level, to ensure that the training, validation and test datasets all have similar species distributions. In total, there are 34,225 training images, 2,679 validation images and 9,565 test images.

\subsection{Dataset Preprocessing}
We preprocessed the dataset provided by New York Botanical Garden before publishing to FGVC competition. These steps include 1) blurring text and barcodes in the image, 2) providing a down-sampled dataset for convenience. 

\subsubsection{Image Blurring}
\begin{figure}[t]
\begin{center}
   \includegraphics[width=0.8\linewidth]{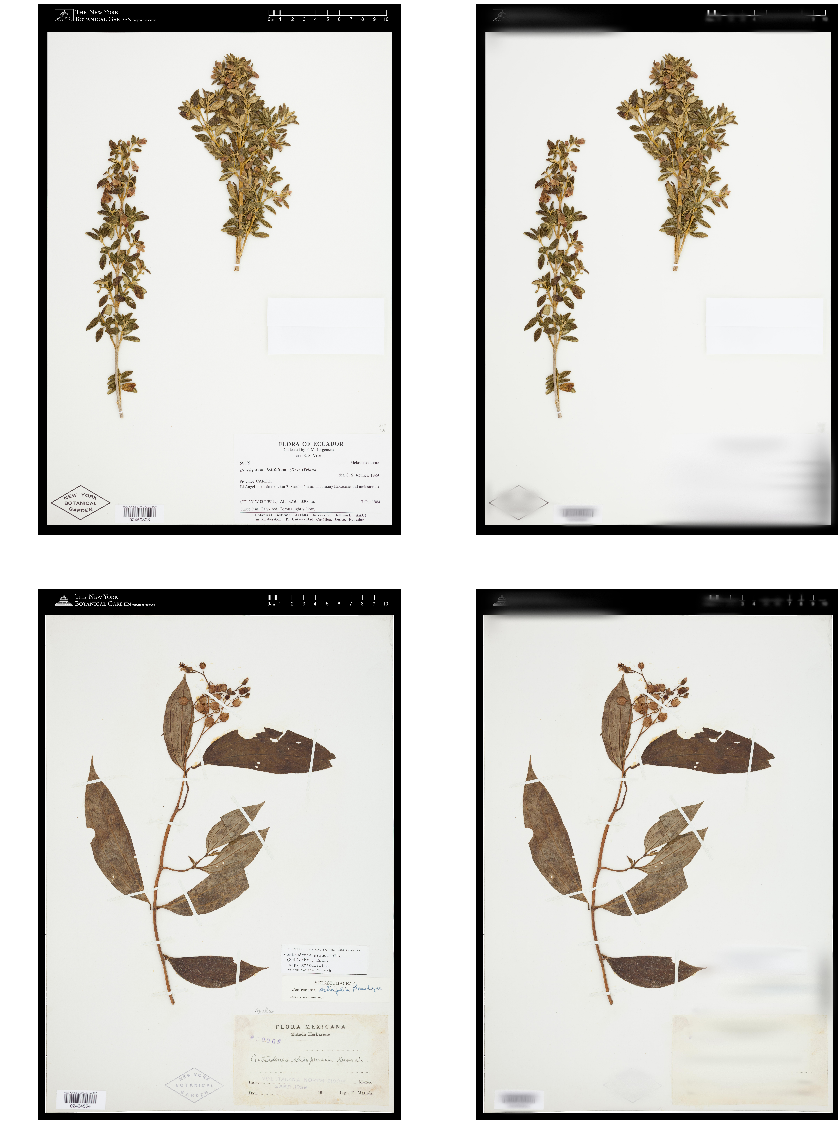}
\end{center}
   \caption{Comparison between original images and blurred images. Left column are original images and right column are blurred images.}
\label{fig:blurred_images}
\end{figure}

The original herbarium images often have handwritten or printed labels attached to it, including date, place found, description of the specimen, and the specimen name. Some sheets also have attached barcodes and identifiers which can be used to trace back to the exact specimen on New York Botanical Garden's website. In order to prevent models from classifying images based on barcode and text information, we have artificially blurred these information (figure \ref{fig:blurred_images}).

The PhotoOCR system \cite{photoocr} was used for text detection. After extracting bounding boxes of text and barcodes, a Heavy Gaussian Blend algorithm was used to blur the detected regions. The algorithm applies a mean blur on the bounding box first, then blurs the regions using a single Gaussian blur with noise added and use a smooth alpha map to blend into original. 

\subsubsection{Image Resizing}
The images collected by the New York Botanical Garden have high resolution, with the majority of images of dimension 6,000 by 4,000. While high resolution images are useful for botanical research, they may not convenient for competition participants to download, process, and train models on.

For convenience, we provided a resized dataset, where each image has been resized (preserving aspect ratios) to have a maximum of 1024 pixels in the larger dimension. This resizing drastically reduced the overall dataset size from 52GB to 2.3GB.

\section{The Herbarium Challenge 2019}
\begin{figure}[t]
\begin{center}
   \includegraphics[width=0.8\linewidth]{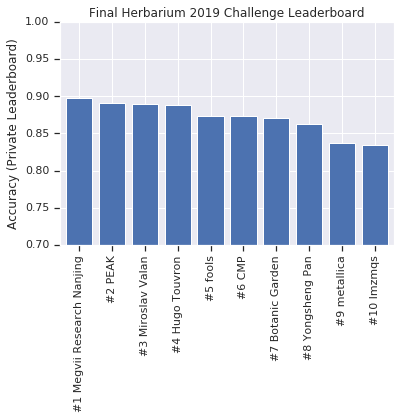}
\end{center}
   \caption{Final challenge leaderboard.}
\label{fig:leaderboard}
\end{figure}

The Herbarium Challenge 2019 was conducted through Kaggle as part of FGVC6 at CVPR19, with 22 participating teams and 254 submissions. The final leaderboard from the held-out private test data can be seen in Figure \ref{fig:leaderboard}. The winning method by Megvii Research Nanjing achieved a classification accuracy of 89.8\%.

Their method involved an ensemble of 5 different models, using the SeResNext-50, SeResNext-101 \cite{XieGDTH16} and ResNet-152 \cite{SENet} architectures. They also used a combination of different losses: cross-entropy, focal loss, and class-balanced focal loss \cite{class-balanced-loss}. Their models were pretrained on the ImageNet \cite{ImageNet} and iNaturalist Challenge 2018 datasets, with input image dimensions of 448x448 and standard data augmentation. Additional techniques used include deformable convolutions \cite{DaiQXLZHW17}, iSQRT \cite{DBLP:journals/corr/abs-1712-01034} and random erasing \cite{DBLP:journals/corr/abs-1708-04896}.

\section{Conclusions and Future Work}
We have developed the Herbarium Challenge dataset to facilitate the development of automatic species identification models for the flowering plant family Melastomataceae. This offers the immediate benefit of accelerating the identification of unlabeled specimens in this family, potentially resulting in the identification of new species.

In future work, we have multiple directions in mind: extending the dataset to incorporate other plant families, adding annotations beyond species, and facilitating new species identification using a test set with previously unseen classes.

\textbf{Acknowledgements.}
We thank NYBG for support and our colleagues at NYBG, particularly, Lawrence Kelly, Damon Little, and Barbara Thiers as well as Nicole Tiernan and Kim Watson for digitization. A special thanks to our NYBG colleague Fabián A. Michelangeli and his collaborators Frank Almeda, Eldis Becquer, Renato Goldenberg, and Walter Judd who explore, describe, and determine melastome species. The digitization of the melastomes in this dataset was made possible by funding from NSF grants including DEB-1343612 and DEB-0818399 to F.A.M. and DBI-0749751, DBI-9987500, DBI-0543335 DBI-1053290 to B.T as well as funding from the Google, Mellon, and Sloan Foundations to NYBG. 

We would also like to thank: Srikanth Belwadi, R.V. Guha and Kiran Panesar from dataCommons.org for help with preparing the dataset; Christine Kaeser-Chen and Hartwig Adam from Google Research; Maggie Demkin from Kaggle; the Herbarium Challenge 2019 competitors, and the FGVC2019 workshop organizers.

\end{document}